# Structured Access

## An Emerging Paradigm for Safe AI Deployment

Toby Shevlane

*University of Oxford*

**Abstract:** Structured access is an emerging paradigm for the safe deployment of artificial intelligence (AI). Instead of openly disseminating AI systems, developers facilitate controlled, arm's length interactions with their AI systems. The aim is to prevent dangerous AI capabilities from being widely accessible, whilst preserving access to AI capabilities that can be used safely. The developer must both restrict how the AI system can be used and prevent the user from circumventing these restrictions through modification or reverse engineering of the AI system. Structured access is most effective when implemented through cloud-based AI services, rather than disseminating AI software that runs locally on users' hardware. Cloud-based interfaces provide the AI developer greater scope for controlling how the AI system is used, and for protecting against unauthorized modifications to the system's design. This chapter expands the discussion of "publication norms" in the AI community, which to date has focused on the question of how the informational content of AI research projects should be disseminated (e.g., code and models). Although this is an important question, there are limits to what can be achieved through the control of information flows. Structured access views AI software not only as information that can be shared but also as a tool with which users can have arm's length interactions. There are early examples of structured access being practiced by AI developers, but there is much room for further development, both in the functionality of cloud-based interfaces and in the wider institutional framework.

# Introduction

The development of new AI capabilities often brings discussion about whether certain uses of AI should be placed off limits. We see this, for example, with facial recognition technology, where a number of U.S. cities have banned the use of the technology by law enforcement. The companies and research groups developing AI ("AI developers") also have a role to play in shaping how the technology is used. AI developers can engage in AI governance through the software they build and how they choose to deploy that software.

Both the field of AI safety, and the recent initiative of AI "publication norms," provide guidance for how developers can shape the impact of the AI systems they deploy. Research in AI safety, broadly defined, offers insights for building AI systems that have beneficial properties: aligned with human values, responsive to human oversight and control, and difficult to misuse for harmful purposes (Christiano, 2016). On the other hand, the recent initiative of "publication norms for responsible AI" (see, for example, Partnership on AI, 2021) concerns the "publication" phase of AI research and development, where research outputs, including trained AI systems, are disseminated. AI developers are encouraged to exercise caution before sharing AI software (and other research outputs) that might have a harmful effect on the world.

These two projects are much more tightly linked than has hitherto been acknowledged. The utility of building safe AI systems is greatly reduced if those systems are then shared in a manner that allows actors to circumvent the relevant safety features. There are many reasons why actors might rework AI systems to be more capable of causing harm, including malicious intent (Brundage et al., 2018), a negligent disregard for risks (Critch & Krueger, 2020), and structural pressures such as economic or geopolitical competition (Bostrom, 2014; Dafoe, 2015). AI is "dual use" on two levels: an AI system can be directly used for both beneficial and harmful purposes, but additionally, AI systems designed to only serve beneficial purposes can be modified to serve harmful purposes. AI developers, therefore, must simultaneously address both how their AI systems can be directly used *and* the pathways through which their AI systems can be adapted and built upon.

This chapter introduces an emerging paradigm for AI deployment, which I refer to as *structured access*. Structured access involves constructing, through technical and often bureaucratic means, a controlled interaction between an AI system and its user. The interaction is structured to both (a) prevent the user from using the system in a harmful way, whether intentional or unintentional, and (b) prevent the user from circumventing those restrictions by modifying or reproducing the system.

The deployment of OpenAI's GPT-3 models (Brown et al., 2020) serves as an early example of structured access. Approved users can access the models through a web interface and an application programming interface (API); external developers can build applications that integrate GPT-3 models via the API, subject to approval. Although this arrangement was partly motivated by commercial factors, it was also motivated by a concern that the models could, if open sourced, be used in harmful ways, such as the mass production of disinformation online (OpenAI, 2020). OpenAI's API platform limits the ways in which the GPT-3 models can be used, and developers who want to build applications using the API must conform to certain safety standards. Another example is Google Cloud's Vision API, which has facial recognition capabilities that only work on a particular set of celebrities, with Google limiting who can use this feature.

Both these examples involve models that are stored in the cloud. Alternatively, it is possible to build structured access principles into models that are disseminated and run on users' hardware. This would involve engineering the model to both conform to certain safety standards and be robust against unauthorized modification. However, cloud-based deployment of AI systems is a much more natural fit with structured access. When the AI system is cloud-based, the developer has many more opportunities to put in place mechanisms for shaping how the system is used. These can be built into the AI system itself, but they can also be built into the interface by which users and application developers interact with the system, and the rules governing who can use the system and for what purposes. These protections can also evolve over time, including revoking any capabilities that turned out to be unsafe—something that is much harder if the underlying software has been disseminated. In addition, cloud-based structured access provides greater security against unauthorized modification of AI systems because the user has no local copy of the software to modify.

Most thinking under the "publication norms" banner assumes that AI research outputs (e.g., models, code, descriptions of new methods) will be shared as information and asks whether certain pieces of information should be withheld. Structured access goes one step further. Structured access leverages the fact that AI models are simultaneously both (a) informational (i.e., exiting as files that can be inspected, copied, and shared) and (b) tools that can be applied for practical purposes. This means that there are two fundamental channels through which AI developers can share their work. They can disseminate the software as information (e.g., by uploading models to GitHub). But they can also lend access to the software's capabilities, analogous to how a keycard grants access to certain rooms of a building. Structured access involves giving the user access to the tool, without giving them enough information to create a modified version. This gives the developer (or those regulating the developer) much greater control over how the software is used, going beyond what can be achieved through the selective disclosure of information.

Therefore, this chapter attempts a reframing of the "publication norms" agenda, adding a greater emphasis on arm's length, controlled interactions with AI systems. The central question should not be: *what information should be shared, and what concealed?* (although that is still an important sub-question). Rather, the central question should be: *how should access to this technology be structured?*

Structured access has not yet reached full maturity and, as we will see, there is much room for continued technical and institutional development. Structured access should not be equated with a lack of transparency over the technology: researchers should explore methods for rendering AI systems transparent (in the ways necessary to provide accountability over safety) without leaking too much intellectual property. Nor should structured access be equated with a centralization of power in the hands of technology companies. Decisions over access could be outsourced to other actors or regulated by governments.

I focus on structured access to trained, deep learning models. Nonetheless, the same principles could be extended to cover the software and hardware used to train these models.[1] I do not discuss legal methods for controlling access to AI systems, such as licenses specifying how a model should be used, but these could be used in tandem with the methods discussed.

The rest of the chapter is organized as follows. First, I introduce the concept of structured access. Then, I will draw the distinction between sharing information and granting access to a tool and argues that there are limits to what can be achieved through the selective disclosure of information. Next, I describe how structured access could apply to both locally disseminated software and cloud-based services but argues that the cloud-based option is superior. Finally, I address the criticism that structured access centralizes power in the hands of AI developers.

## What Is Structured Access?

Structured access is a paradigm for safe AI deployment. The aim is to prevent an AI system from being used harmfully, whether the harm is intended by the user or not. The developer offers a controlled interaction with the AI system's capabilities, using technical and sometimes bureaucratic methods to limit how the software can be used, modified, and reproduced. The methods of structured access contrast with the distribution of open-source software in that structured access places greater distance between AI capabilities and the actors accessing those capabilities.

Structured access is rooted in a broader phenomenon, going beyond AI, where the owners of potentially harmful artefacts attempt to place limits on how users can interact with those artefacts. For example, certain biological laboratories have the capability to print DNA sequences and offer this as a service. The synthesized DNA can be used for beneficial research but could in theory be used for the creation of bioweapons. This means that the lab must screen orders to ensure that the synthesized DNA will be used safely. In addition, because the printing technology is becoming more widely available, there are calls for DNA synthesis machines to be manufactured with hardware-level "locks" on what they can print (Esvelt, 2018). Both these methods work by adding structure to the interaction between the actor requesting the synthesis and the underlying printing capability.

If structured access means building a controlled interaction between the AI system and the user, then what exactly is being controlled? There are two broad categories: (1) *use controls*, which govern the direct use of the AI system (who, what, when, where, why, how?); and (2)

*modification and reproduction controls*, which prevent the user from altering the AI system or building their own version in a way that circumvents the use controls.

## Use controls

The purpose of the use controls is to ensure that the AI system is used in a safe and ethical way. The controls govern who can access the AI system and for what purpose, the underlying technical features of the AI system, and the scope of capabilities on offer. Partly this involves careful design of the AI system itself, taking into account the system's safety, its alignment with human values, whether its outputs are fair and unbiased, and whether its capabilities could be misused by actors seeking to intentionally cause harm. These are "software-level use controls." But the use controls need not be built into the AI system itself—they can alternatively be built into the procedures through which the user accesses the AI system. These are "procedural use controls." For example, if a developer wants to avoid their language model being used for medical diagnosis, then rather than, say, removing medicine-related text from the model's training data, the developer could vet users and deny access to anyone hoping to use the model for diagnosis.

## Modification and reproduction controls

Let's imagine that the developer has built an AI system to a high standard of safety, putting in place many software-level use controls. The developer then open sources that system, including public sharing of the model's weights, and any code, datasets, and simulated environments used to train the system. The open-source approach benefits researchers and product developers who can now build upon the system, make modifications, and extend the work. Indeed, modifiability is a central tenet of open source software (Kelty, 2008). However, the same modifiability enables the user to remove software-level use controls (e.g., by retraining the model such that it learns a capability that the developer had purposefully avoided). Similarly, the open-source approach makes the AI system reproducible—users can re-run the training regime themselves. This means that they can make changes to the training regime, such as changing the data on which the system was trained or editing the loss function such that it no longer penalizes unsafe behavior. Additionally, even if the developer puts in place procedural use controls on who can access the AI system and for what purpose, if the AI system is easily reproducible, other actors will be able to share the system without these controls.

Therefore, to successfully protect the use controls against removal or circumvention, the developer must also introduce controls on modification and reproduction of the AI system. There are certain obvious, preliminary methods for doing so. The developer can decline to open source the model and the training code. The developer can put in place strong cybersecurity defenses against theft of such information from its servers. However, below we will see more sophisticated modification and reproduction controls. For example, for models that run on users' hardware, there are methods for making the model immune to fine-tuning. For cloud-based AI services, the developer can go one step further and allow the user to, at arm's length, make authorized (but not unauthorized) modifications to the model.

## The Limits of Selective Information Disclosure

Structured access is related to the concept of structured transparency (Bostrom, 2019; Shevlane et al., 2020; Trask et al., 2020). According to Trask et al. (2020, p. 3), "A system exhibits structured transparency to the extent that it conforms with a set of predetermined standards about who should be able to know what, when they should be able to know it, and what they should able to do with this knowledge; in other words, if it enforces a desired information flow." This involves finding mechanisms, both technical and social, for granting access to certain pieces of information while keeping others private (the archetypal example being a sniffer dog, which signals if a bag contains explosives without revealing the other contents of the bag). Structured access differs in that, where structured transparency would govern what information somebody knows and does not know about an AI system, structured access goes further by governing what somebody can and cannot do with an AI system.

There is some overlap between these two. As we saw above, structured access will normally involve limiting access to information about the AI system's design, as part of the controls on modification and reproduction. Blanket secrecy will usually be unnecessary: some forms of information about an AI system will not be very useful for modification and reproduction, and yet will have high social value. For example, for AI safety reasons, developers could grant academic researchers the ability to, at arm's length, run interpretability tests on an otherwise private model, allowing high-level insights to be gleaned about how the model functions. Such opportunities for selective disclosure of information mean that the techniques of structured transparency discussed by Trask et al. (2020) are relevant and useful within structured access.

However, structured access involves more than just controlling the flow of information about AI systems. A file containing the parameters of a deep learning model is not only a piece of information, but also something that, when combined with certain computing infrastructure, can be used as a tool. As philosophers of software have pointed out, software code has a dual nature, as both a piece of text, written in a formal language, which can be read and interpreted, but also as something that functions in a machine-like way (Colburn, 1999). As information, the weights of a language model can be inspected, copied, and shared; but on the right computer, they can also be used for practical purposes, for example, to write the conclusion of an essay, or to classify social media posts into political categories.

Structured access involves separating the software-as-a-tool from the software-as-information, such that the user can access the tool without accessing the information necessary to recreate it.[2] This allows the tool to gain widespread use, even if its design has not been disclosed.[3] It also opens a new set of opportunities for the developer to put in place use controls. We will see many examples of this below; for example,  the developer can enforce rules on what queries are fed into the model; the developer can monitor how the model is being used, shutting down users who violate the rules; or the developer can retract certain capabilities that, with time, appear harmful.

This level of control is not possible if the developer is forced to only rely on selectively disclosing the informational content of AI software. Selective disclosure of information works well for factual information—such as personal medical data or the number of nuclear warheads that a country possesses—where some facts can be communicated and others kept private. However, this does not work so well for the governance of dual use technologies (e.g., for blueprints for the design of a weapon or software code). Here, the information is of a different character: it is "how to" information, with certain practical applications, and those applications are numerous, some beneficial and others harmful. The fundamental problem is that there exists no sweet spot where the user knows enough to achieve only the beneficial ends. Either they know too little, and so cannot use the technology at all, or they know too much, and so they have too much leeway.

This problem is illustrated by DialoGPT, a chatbot built by researchers at Microsoft (Zhang et al., 2019). Due to concerns that the chatbot's outputs were often "unethical, biased or offensive," the researchers open sourced the model but without the final piece of code necessary for producing text from the model. This is no solution to the mixed nature of the model's outputs because either users cannot get the model to produce any text whatsoever, or they find substitute versions of the missing code online and thereby have access to the full functionality of the model. Perhaps users who are determined enough to do the latter are, on average, more likely to use the model responsibly. But even so, this is a very imprecise and insecure method of controlling how people use the model. The lesson is that the developer cannot selectively filter the informational content of the software in a way that neatly discriminates between uses.

On top of this, there are two additional difficulties with using selective disclosure as a means of controlling how people use the software. First, the donor of information cannot later change their mind and revoke the information that they chose to share. This is a serious problem given the difficulty of forecasting the impact of a technology in advance of deployment and given that the impact of the technology might depend on factors that change over time (such as the presence of complementary technologies, or the propensity of actors to misuse the technology). Second, information is easily passed on. If details about an AI system are disseminated to a specific group of recipients, there is no guarantee that they will not disseminate that information further. Different actors will be more or less prone to misusing AI technologies, and so the developer may wish to exercise high levels of control over who can use a particular system. In contrast, if the developer deploys the AI system through an API or web interface, they can require users to log into the service.

## Implementation of Structured Access: Local and Cloud-based Systems

AI developers, like software makers more generally, have two broad options for deployment. In one case, the user is given a local copy of the software that they can run on their own hardware.

In the other case, the user has no local copy, and interacts with the model remotely through an interface; the model is stored on an external server. The latter approach reflects the broad trend towards "software as a service" (and related phenomena such as "platform as a service"), whereby today, users commonly interact with software through interfaces such as web browsers (e.g., Facebook, Google Docs), without necessarily downloading the software onto their computers.[4]

AI developers currently rely on both these modes of deployment. It is common for research groups, including those within large AI companies, to publicly share copies of their deep learning models on sites such as GitHub, where the models are free to download. Another example is that, where AI software is used on smartphones (such as augmented reality filters for videos), a local copy of the software will often be stored on the phone. On the other hand, AI software is also widely built into the backend of cloud services, such as Google Search or Grammarly (which uses deep learning models stored in the cloud to correct users' grammar). Some AI developers also bundle API model access together with their cloud computing services.

Structured access can be implemented on both local and cloud-based systems. However, we will see that cloud deployment is a better home for structured access, offering greater opportunities for controlling what the user does with the AI system. The cloud-based approach should be viewed as the archetypal way of implementing structured access.

## Local deployment

Under the local deployment approach to structured access, the AI system runs on the user's hardware. The AI system will contain software-level use controls, as well as software-level protections against unauthorized modification and reproduction (such as obfuscation of the model). The result of these controls will be that, although the model sits on the user's hardware, the user has only a constrained interaction with the model.

## Use controls

Where the developer distributes local copies of AI software, the obvious place for use controls is at the software level. In the case of deep learning models, this will often involve careful engineering of the model itself. Alternatively, because AI systems often combine multiple deep learning models and other algorithms, the safety mechanism could be more modular; for example, a language model for text generation could be used alongside a model that classifies outputs and filters out certain kinds of statements.[5]

One approach is to construct the training regime such that certain dangerous capabilities are never learned by the model. A rudimentary example occurred with OpenAI's GPT-2 models. The developers trained a range of model sizes, initially only sharing smaller versions, believing these smaller models to be less capable of misuse (e.g., producing less convincing fake news articles) (Solaiman et al., 2019). However, in this case, it was not only the risky capabilities that were weaker—the smaller models were generally less performant. Indeed, manipulating model size is a very imprecise method for controlling how the model is used. Because of this, as well the lack of controls on modification, GPT-2 should not be considered an example of structured access—it is an example of selective disclosure.

The developer can also put in place procedural use controls, although the scope for doing so is greatly reduced in the case of local deployment. For example, instead of publicly sharing the software, the developer could decide to disseminate the software to only a select group of users. Moreover, the developer could use a software license key system, where the software is only usable once a valid key has been entered. Such a system could in theory help to limit the range of users who can access the software, and even allow the developer to revoke access to the software (if the license is time-limited). However, these methods are likely to be less secure than the equivalent in cloud-based deployment. In addition, unlike the cloud-based approach, there is no way for the developer to directly monitor and police how the software is being used.

## Modification and reproduction controls

If the user has a local copy of the model, one threat is that the user will make modifications to the model, for example, by fine-tuning on a new task, or going into the code and removing features that were installed for safety reasons. Another threat is that the user gleans reproduction-relevant information from the model. This could be done through inspecting its architecture or copying elements of the model, such as the weights of certain layers of the network. Alternatively, the user could train a new model using the outputs of the existing model as data; that is, model stealing (Tramèr et al., 2016).

A small body of literature offers technical methods for securing deep learning systems against unauthorized modification and reproduction. The main themes of the literature are preventing fine-tuning and making the neural network less transparent (e.g., via obfuscation or encryption). Three methods are outlined in Table 39.1.

| Method | Description |
|--------|-------------|
| "Deepobfuscation" via knowledge distillation (Xu et al., 2018) | The model is distilled (i.e., a new model is trained on its outputs) into one that is smaller and has a simpler structure. This conceals the structure of the initial model; and the new, smaller model less well-suited to fine-tuning. |
| "MimosaNet": Increased sensitivity to weight changes (Szentannai et al., 2019). | Fine-tuning is prevented by making the model "extremely sensitive" to changes in the model weights. This is achieved through adding additional weights to |

| | the model which are designed to leave the performance of the model unchanged, yet which dramatically alter the performance of the model if they are modified. |
|---|---|
| Encryption of the model weights (Alam et al., 2020; Chakraborty et al., 2020; Xue et al., 2021) | The model is trained such that the input of a secret key is necessary for it to perform well. The secret key can also be stored on trusted hardware. This allows the developer to limit who can use the model. In addition, the encrypted weights are resistant to fine-tuning. |

*Table 39.1: Technical innovations in the protection of deep learning models against modification and reproduction.*

These methods are specific to deep learning models. Generally applicable methods of protecting software against piracy are also relevant. For example, one technique would be to compile the model into binary form before sharing it with users. (Nonetheless, there is no guarantee that a well-motivated attacker could not recover the model, especially if tools for decompiling become more sophisticated.) Another approach is to embed the software within hardware that is difficult for the user to access, as can be done with autonomous vehicles.

Generally, applying structured access to locally deployed AI systems is an uphill struggle because it is harder to control how the user interacts with software that is running on their own computer. Nonetheless, developers may sometimes be forced to rely on the local deployment approach where the cloud-based approach is unavailable. For example, in the case of autonomous vehicles, the system cannot rely on always having a strong internet connection. Such cases motivate further technical innovation in structured access for local deployment.

## Cloud-based deployment

The second implementation is where the user never gains possession of the model, but instead interacts with the model through an interface. This is the strategy adopted by OpenAI for their

GPT-3 models, where users are given an API and a web interface. Strictly speaking, this category of deployment approaches need not always involve a cloud service. For example, DeepMind initially disclosed to biologists predictions about the structure of COVID-19 from their AlphaFold 2 model (Jumper et al., 2020),[6] without building a cloud interface for biologists to use. The essence is that the model is stored and run on hardware that is not in the possession of the users.

## Use controls

As with local deployment, there is ample room for software-level use controls. The developer could then grant the user access to the full functionality of the model through an interface. However, cloud-based deployment excels in the area of procedural use controls.

Instead of having to encode distinctions between different uses into the software itself, the developer can set enforceable rules about what users can do with the model. For example, with GPT-3, OpenAI has different categories of uses: those which are considered generally safe (e.g., extracting keywords from a passage); those which are disallowed (e.g., using GPT-3 for spam); and those which will be evaluated on a case-by-case basis, depending on the sensitivity of the domain and other considerations (e.g., using GPT-3 for social media posts). Compliance with such rules can be enforced because the developer can monitor how users are interacting with the model and block access from users who violate the rules. (Monitoring can be done in a privacy-preserving way; see Trask et al., 2020.) The developer can also change the rules over time.

Generally, the developer can create a highly mediated interaction between the model and the user, with multiple layers of use controls in place. As well as rules that separate between different kinds of uses, the developer can also separate between different kinds of user. For example, the developer could only grant access to actors who are unlikely to misuse the model or could offer tiered access for different types of actors. The developer can also impose rate limits or quotas on the number of times that each user can query the model.

There may also be a layer where external developers build applications on top of the model, with end users interacting with those applications. With OpenAI's GPT-3 models, this arrangement helps with refining the range of possible uses of GPT-3 because the applications are bounded in scope. For example, OpenAI is keen to avoid downstream developers from building an open-ended text generator or chatbot using GPT-3, whereas their documentation clarifies that a chatbot constrained to only answering mathematics questions would likely be acceptable (OpenAI, 2021).

An interesting development is that, with GPT-3, the application developer can use "prompt engineering" to achieve this narrow focus. For example, the application's code could insert the user's question into a standardized template submitted to the API, such as the following:

Question: What is the term for the long side of a triangle?

Answer: The hypotenuse.

Question: What is the mathematical term for a whole number?

Answer: An integer.

Question: Who was the Democratic candidate in the 1996 U.S. presidential election?

Answer: Sorry, I only answer questions about mathematics.

Question: **[Insert user's question]**

Answer:

GPT-3's continuation can then be submitted to the user as the response from the question-answering chatbot. The hope is that the model will generalize from the examples given in the prompt and only answer questions about mathematics. The application developers can experiment with different prompt templates to find one that produces the desired kinds of answers. Moreover, the API provider can monitor what prompts are being fed into the API, and so in theory could notice if the application developers try to make the chatbot more open-ended than was previously promised. This is a neat illustration of how AI developers, using cloud-based structured access, are able to implement a more holistic approach to safety, manipulating not only the model's internal makeup but also the process through which the user accesses the model's capabilities.

## Modification and reproduction controls

In the extreme case, the user has no access to the inner workings of the model. The user therefore cannot directly modify the model, nor inspect it for the purposes of recreating a modified version. In addition, the developer can install checks against model stealing (i.e., where users train a new model by using the original model's outputs as data), such as by imposing quotas on the number of times the user can query the model.

Nonetheless, the cloud-based approach is sufficiently flexible to accommodate controlled modification and inspection of the model. For example, at the time of writing, OpenAI plans to allow for fine-tuning of its GPT-3 model. Modification could be done at arm's length, with developer vetting proposed modifications on safety and ethical grounds.

Developers could also build functionality for external researchers to subject the model to interpretability analyses, as a way of facilitating study of the model's safety. An interesting question is how much depth of analysis could be permitted without giving away too much knowledge relevant for recreating the model (e.g., could researchers study specific neurons, or only the more high-level functioning of the model, such as the activations of the different layers?). One arrangement for preserving the privacy of the model could be that the external researcher uploads to the developer a tool that automates the interpretability analysis, then the developer runs that tool on the model and sends back the results. Another option is that the cloud-based interface grants deeper model access to approved, trustworthy researchers, who agree not to share reproduction-relevant information about the model. Overall, improvements in the functionality of these interfaces should allow cloud-based structured access to recoup some of the advantages of open source software (i.e., modifiability and interpretability).

Finally, one problem with the cloud-based approach is that it currently requires the developer to invest in the supporting infrastructure, which might be too burdensome for lower-resource developers of AI systems, including many academic labs. One possibility to be explored is the creation of an organization that acts as a centralized hub for cloud-based structured access; that is, the structured access equivalent of GitHub or Hugging Face. This platform could host many different models from many different developers and employ experts to make standardized and externally informed decisions about what capabilities should be accessible to whom.

## The Centralization of Power

An obvious criticism of the vision of structured access presented so far is that it centralizes power in the hands of AI developers. One concern might be that the developers cannot be trusted to make the right decisions about what capabilities should be granted to whom, especially if they have a commercial interest in selling AI services. A related concern is the possible lack of accountability in how AI developers make these decisions.

The first step to mitigating these concerns is acknowledging that structured access can be embedded in a wider governance framework. For example, when governments regulate the use of AI, they could require AI developers to avoid undermining those regulations when providing access to AI systems. This would be analogous to the regulation of underage drinking, where governments ban retailers from selling alcohol to underage individuals. In addition, governments could regulate AI deployment in a more hands-on way, setting standards for what capabilities should be made available, in what form, to whom. In tandem, governments could require that AI developers facilitate some minimum level of access by external researchers who perform interpretability tests on the relevant AI system. Alternatively, these standards could be set by independent civil society bodies.

We can also cast doubt on the pervasive assumption that says widely distributing AI knowledge will, accordingly, widely distribute power. There are two reasons to doubt this. First, if users gain unencumbered access to the technology, what has been distributed? The power to set communal standards over use of the technology has not been distributed to users. In Hohfeldian terms, the users are each granted a *freedom*; that is, a freedom to use the technology how they like, rather than a "power" over how others use the technology (see Hohfeld, 1919). The only way to actually distribute the power that structured access (de facto) gives to the developer is to embed structured access in a wider governance framework. Another way of putting the same point: there are two different definitions of AI "democratization."[7] One definition is simply that a wide range of people can use the technology. The other definition is more faithful to the meaning of "democracy" in the political arena and asks whether users can collectively influence how the technology is designed and deployed. These two types of democratization sometimes conflict: if AI developers open source their systems, this grants wide access to the system, but simultaneously undermines attempts to build communal decision-making structures within AI deployment.

Second, we can ask: what are the downstream consequences, in terms of the distribution of power in society, of all actors gaining unencumbered access to AI technologies? For many AI technologies, there are reasons to doubt that actors with low levels of existing power would be the primary beneficiaries. As a crude analogy, consider what happens when modern handcuffs become widely available. Handcuffs assist law enforcement officers in making arrests because the officers are legally permitted to use them—it is not that the power to make arrests becomes more evenly distributed. Likewise, there are many AI technologies that will likely increase the power of actors with existing power resources, such as legal authority, data collection abilities, and financial resources (see Pasquale, 2016). For example, an employee cannot come into work one morning and ask to use an AI-powered worker surveillance system on their boss. This is a complex issue, and deserves much more in-depth discussion (see Shevlane & Dafoe, 2021). But suffice to say, there is no simple relationship that links wider availability of AI systems with an equalization of power in society. In fact, structured access could be used for preventing certain actors from abusing their power—for example, by denying access to governments who wish to use AI for political repression.

## Conclusion

The theory presented in this chapter should be integrated into an updated understanding of the "publication norms" agenda within AI governance. The publication norms agenda is defined by the Partnership on AI, for example, as "the consideration of *when and how to publish novel research* in a way that maximizes benefits while mitigating potential harms" (Partnership on AI, 2021, emphasis added). In discussions about publication norms, the focus is often on the developer's ability to make certain forms of information public (i.e., research outputs, and discussions about risk), whilst potentially carving out certain research outputs to be kept private. Going forward, AI governance must focus on *both* the sharing (and withholding) of information about AI systems *and* the growing infrastructure for hosting arm's length interactions between users and AI systems.

The early examples of structured access are encouraging, but there is much room for improvement. Two areas should be the focus. First, there is the functionality of cloud-based interfaces. As I argued above, these interfaces could grant users and researchers greater access to the model whilst still maintaining robust controls on modification and reproduction. The user could be given the ability to fine-tune (or otherwise modify) the model in a controlled setting. External researchers could also be given the ability to run interpretability tools on the model. There is also the question of user privacy: how can the privacy of the user's activity be respected, even though the use of the model is being monitored for compliance with safety standards? Generally, if the interface's functionality can be improved without sacrificing on safety, it is important that these improvements are implemented, such that cloud-based structured access can recover some of the benefits of the open-source approach.

Second, there is the overarching framework of governance within which these cloud-based interfaces sit. One possibility, which can be described as "GitHub for structured access," would involve creating a central repository for models. This repository would be overseen by a body that makes standardized, informed decisions about how access to each model should be structured, while also monitoring compliance with these standards. This would benefit research groups and AI companies who do not have the internal capacity for implementing structured access. Another possibility is that structured access is given legal backing, with governments setting regulatory standards for AI deployment. This could increase the legitimacy and uniformity of decisions over access to models. Overall, structured access is in its infancy, and there is much room for technical and institutional development.

## Acknowledgments

I am grateful for helpful comments and suggestions from: Markus Anderljung, Carolyn Ashurst, Ondrej Bajgar, Avital Balwit, Nick Bostrom, Justin Bullock, Allan Dafoe, Eric Drexler, Ben Garfinkel, David Krueger, Jeffry Ladish, Alex Lintz, Luke Muehlhauser, Laurin Weissinger, and Baobao Zhang.

---

[1] For example, when developers buy cloud compute for training a model, the provider could check that the proposed model conforms with certain safety standards. Brundage et al. (2020, p. 32): "Cloud providers already employ a variety of mechanisms to minimize

risks of misuse on their platforms, including "Know Your Customer" services and Acceptable Use Policies. These mechanisms could be extended to cover AI misuse. Additional mechanisms could be developed such as a forum where cloud providers can share best-practices about detecting and responding to misuse and abuse of AI through their services."

[2] Note that it is often impossible to completely isolate the software-as-a-tool such that no information about the workings of the tool leaks out during use. If an alien came to earth and encountered our cars for the first time, they could glean some insights about how the car works simply by driving one around, without needing to see any blueprints. In the same way, somebody interacting with an AI system through a web interface could pick up on certain reproduction-relevant information. One simple insight is that the AI capabilities on display are actually achievable. They could also consider the form that the inputs to the model must take (for example, GPT-3 is an autoregressive language model, and, as a consequence, it will predict how a passage of text might be continued, rather than filling in words in the middle of a passage).

[3] This vision of AI deployment is highly compatible with Drexler's (2019) vision of "comprehensive AI services," where advanced AI systems are deployed in service of bounded tasks across the economy.

[4] Note that many services, especially smartphone apps, rely on a mixture of locally downloaded software and cloud-based software.

[5] That said, modular features might be easier to remove through modification.